\documentclass[conference]{IEEEtran}
\IEEEoverridecommandlockouts

\usepackage{fancyhdr}
\usepackage{cite}
\usepackage{amsmath,amssymb,amsfonts}
\usepackage{algorithmic}
\usepackage{graphicx}
\usepackage{url}
\usepackage{textcomp}
\usepackage{float}
\usepackage{xcolor}
\usepackage{comment}
\usepackage{subcaption}
\usepackage{multirow}
\usepackage{makecell}

\def\BibTeX{{\rm B\kern-.05em{\sc i\kern-.025em b}\kern-.08em
    T\kern-.1667em\lower.7ex\hbox{E}\kern-.125emX}}
\title{Optimized Weighted Voting System for Brain Tumor Classification Using MRI Images\\
}

\author{\IEEEauthorblockN{1\textsuperscript{st} Ha Anh Vu}
\IEEEauthorblockA{\textit{Artificial Intelligence Department} \\
\textit{FPT University}\\
Ho Chi Minh City, Vietnam \\
{0009-0002-4921-9823}} 
} 
\begin{document}

\maketitle
\begin{abstract}
The accurate classification of brain tumors from MRI scans is essential for effective diagnosis and treatment planning. This paper presents a weighted ensemble learning approach that combines deep learning and traditional machine learning models to improve classification performance. The proposed system integrates multiple classifiers, including ResNet101, DenseNet121, Xception, CNN-MRI, and ResNet50 with edge-enhanced images, SVM, and KNN with HOG features. A weighted voting mechanism assigns higher influence to models with better individual accuracy, ensuring robust decision-making. Image processing techniques such as Balance Contrast Enhancement, K-means clustering, and Canny edge detection are applied to enhance feature extraction. Experimental evaluations on the Figshare and Kaggle MRI datasets demonstrate that the proposed method achieves state-of-the-art accuracy, outperforming existing models. These findings highlight the potential of ensemble-based learning for improving brain tumor classification, offering a reliable and scalable framework for medical image analysis.
\end{abstract}

\section{Introduction}
\label{introduction}
Magnetic Resonance Imaging (MRI) is a critical tool for detecting and diagnosing brain tumors, where accurate classification is essential for guiding treatment decisions. Tumors such as glioma, meningioma, and pituitary tumors exhibit diverse morphological characteristics, including variations in shape, texture, and anatomical location. These differences and variations in MRI acquisition conditions pose challenges for precise classification. Traditional machine learning models like Support Vector Machines (SVM) and K-Nearest Neighbors (KNN) have shown effectiveness when leveraging handcrafted features such as Histogram of Oriented Gradients (HOG). Meanwhile, deep learning models, particularly Convolutional Neural Networks (CNNs) and ResNet architectures, have demonstrated promising results in medical imaging tasks. However, both approaches have limitations when applied independently, necessitating an integrated strategy to enhance classification performance.

This research proposes a weighted ensemble learning framework that combines multiple classifiers to improve brain tumor classification accuracy and robustness. The ensemble integrates traditional machine learning models with deep learning architectures, leveraging their complementary strengths. A weighted voting mechanism is employed to optimally combine predictions from different classifiers based on their individual performance. Image processing techniques such as Balance Contrast Enhancement (BCET), K-means clustering, and Canny edge detection are applied to enhance feature extraction, improving tumor boundary visibility and classification accuracy. Specifically, SVM and KNN are used for shape and texture analysis, while CNN-MRI, ResNet101, and DenseNet121 process grayscale MRI images to extract deep spatial representations. ResNet50 and Xception are also trained on edge-detected images to enhance the extraction of structural features.

Experimental evaluations on the Figshare and Kaggle MRI datasets demonstrate that the proposed ensemble framework achieves classification accuracy exceeding 99\%, surpassing existing methods and highlighting its potential for real-world medical applications. The remainder of this paper is structured as follows: Section \ref{relatedwork} reviews related work on brain tumor classification using single and ensemble models. Section \ref{methodology} details the methodology, including image processing, model selection, and the implementation of the weighted voting mechanism. Section \ref{results} presents the classification results and compares them with state-of-the-art approaches. Finally, Section \ref{conclusion} concludes the paper with key findings and directions for future research.

\vspace{-0.15cm}
\section{Related Work}
\label{relatedwork}
Advancements in brain tumor classification have been driven by both single-classifier models and ensemble-based approaches. While individual classifiers, such as deep learning-based Convolutional Neural Networks (CNNs) and traditional machine learning models like Support Vector Machines (SVM) and K-Nearest Neighbors (KNN), have demonstrated strong performance, each has its limitations in handling variations in tumor morphology. Ensemble learning addresses these challenges by integrating multiple models, leveraging their complementary strengths to improve classification accuracy and robustness. This section provides an overview of both single-model and ensemble-based approaches in brain tumor classification.
 
\subsection{Single Classifiers}

CNNs are widely used in medical imaging for their ability to extract hierarchical features from complex datasets. ResNet, VGGNet, and DenseNet have achieved state-of-the-art performance, particularly when fine-tuned with large-scale datasets like ImageNet \cite{res-survey,SWATI201934}. Further improvements, such as hyperparameter tuning and architectural enhancements, have significantly boosted classification accuracy \cite{Khan2020, Rasheed2023, Rasheed2024, Irmak2020}. Recent advancements include sub-region tumor analysis \cite{cheng-brain}, EfficientNet optimization \cite{Ishaq2025}, and the TDA framework for segmentation, classification, and severity prediction \cite{Farhan2025}.

Traditional machine learning models remain relevant, especially in resource-limited settings. SVM \cite{svm} and KNN \cite{KNN}, combined with feature extraction techniques like HOG and LBP \cite{HOG}, continue to perform well. Multi-kernel SVM \cite{Dheepak2020, sadad2021} and handcrafted features \cite{knn-figshare, havaei2014efficient} have sometimes rivaled deep learning methods. Hybrid approaches, such as integrating Canny edge detection with CNNs \cite{Canny, Vu, vu2024a}, further enhance classification accuracy by preserving spatial and textural information.

\subsection{Ensemble Learning}

Ensemble learning enhances predictive accuracy by combining multiple classifiers, reducing variance, and improving generalization, making it highly effective in medical imaging. Techniques like bagging, boosting, and voting have been widely explored. In brain tumor classification, Bogacsovics et al. \cite{Bogacsovics2023} used majority voting with CNNs like ResNet and MobileNet, while Siar et al. \cite{Siar2021} applied weighted voting with VGG16 and ResNet50, improving performance. Sterniczuk et al. \cite{multimodels} further validated ensemble methods by integrating multiple deep learning models, leveraging diverse feature extraction techniques for higher accuracy.

Hybrid ensemble models have proven effective in improving accuracy. Munira et al. \cite{Munira} combined CNNs with Random Forest using a voting mechanism, while Zhao et al. \cite{Zhao2021} integrated SVM ensembles with PCA for dimensionality reduction. These approaches demonstrate the benefits of combining classifiers for enhanced performance.

Recent advancements in ensemble learning have incorporated attention mechanisms and advanced feature fusion techniques to refine classification accuracy. Abdulsalomov et al. \cite{Abdulsalomov2023} enhanced model reliability by integrating the Convolutional Block Attention Module (CBAM) and Bi-directional Feature Pyramid Networks (BiFPN) into an ensemble framework. Roy et al. \cite{roy} introduced a GAN-based augmentation technique within an explainable ensemble system, effectively addressing data imbalance issues and improving overall classification performance.

Building on these prior works, this study proposes an enhanced ensemble learning framework that integrates deep and traditional machine learning models. A previous study by Vu et al. \cite{Vu2024} explored an ensemble comprising KNN, SVM, CNN-MRI, and ResNet50, achieving a peak accuracy of 98.36\%. In contrast, the proposed method expands upon this by incorporating additional deep learning models, Xception, ResNet101, and DenseNet121, further improving classification accuracy beyond 99\%. 

\section{METHODOLOGY}
\label{methodology}
This study explores the integration of multiple classifiers with diverse data representations. Each model, when trained on different input types, contributes unique classification perspectives for brain tumors. These variations complement one another within the ensemble system, ultimately improving accuracy. The choice of classifiers and input types is informed by preliminary experiments \cite{Vu, Vu2024} and their proven effectiveness, as outlined in the section \ref{relatedwork}.

\subsection{MRI data representations}

The experiments used various data representations, including original images, edge images, and features like HOG \cite{HOG}, SIFT, ORB, EHD, PCA, color histograms, and LBP. Prior research found HOG, edge images, and original images most effective for MRI processing, leading to their selection for this study.

\subsubsection{Grayscale images}

The grayscale MRI images will be maintained at their native resolution of 512x512 pixels to preserve their structural fidelity and prevent distortions caused by resizing. The Balance Contrast Enhancement Technique (BCET) \cite{bcet} will improve the visibility of tumor components by enhancing contrast, making key features more distinguishable.

\subsubsection{Histogram of Oriented Gradients (HOG)}

HOG \cite{HOG} is a feature extraction method for brain tumor analysis, capturing shape and edge structures while remaining resilient to contrast and illumination changes. Unlike SIFT and ORB, which focus on key points, HOG preserves global structures crucial for MRI-based classification. Color histograms lack spatial details, EHD struggles with irregular tumor shapes, and LBP is noise-sensitive. PCA risks losing essential spatial features. HOG effectively balances global and local structures, preserving tumor boundaries and improving classification accuracy.



\subsubsection{Edge images}
\begin{figure}[!ht]
    \centering
    \begin{subfigure}[b]{0.24\linewidth}
        \centering
        \includegraphics[width=\linewidth]{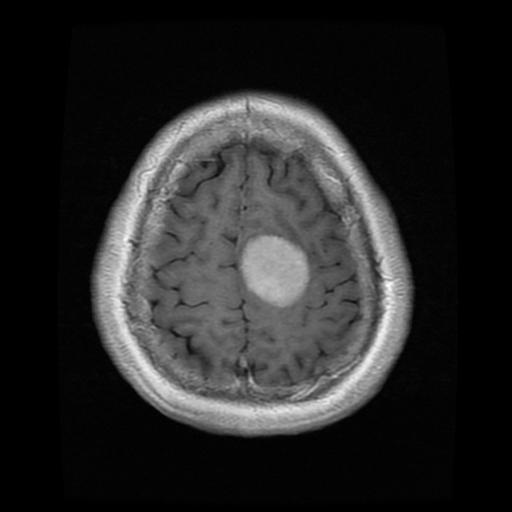}
        \caption{}
        \label{fig:original2}
    \end{subfigure}
    \hfill
    \begin{subfigure}[b]{0.24\linewidth}
        \centering
        \includegraphics[width=\linewidth]{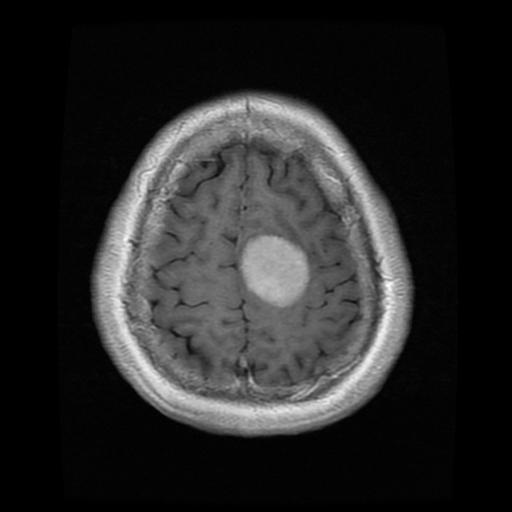}
        \caption{}
        \label{fig:enhanced}
    \end{subfigure}
    \hfill
    \begin{subfigure}[b]{0.24\linewidth}
        \centering
        \includegraphics[width=\linewidth]{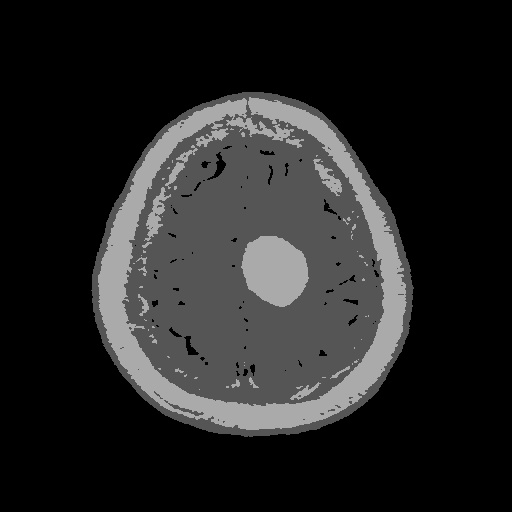}
        \caption{}
        \label{fig:segmented}
    \end{subfigure}
    \hfill
    \begin{subfigure}[b]{0.24\linewidth}
        \centering
        \includegraphics[width=\linewidth]{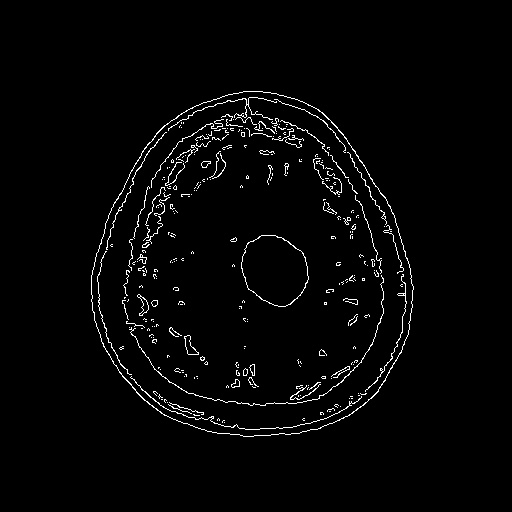}
        \caption{}
        \label{fig:edge}
    \end{subfigure}

    \caption{Edge detection process on MRI image: (a) Original (b) Contrasted (c) Segmented (d) Edge detected.}
    \label{fig:processing_steps}
\end{figure}

Accurate tumor classification relies on well-defined edges to enhance MRI visibility. The pipeline includes grayscale conversion, BCET for contrast enhancement, and K-means clustering to segment the skull, soft tissues, and tumor \cite{Vu}. Canny edge detection \cite{Canny} refines tumor boundaries, using a 5×5 Gaussian filter for smoothing, a 3×3 Sobel operator for gradient detection, and non-maximum suppression to retain prominent edges. Double thresholding and hysteresis edge tracking remove false positives. Figure \ref{fig:processing_steps} illustrates the process, from the original MRI image to contrast enhancement, segmentation, and final edge-detected output, improving classification accuracy.

\subsection{Classifiers}
\subsubsection{K-Nearest Neighbors (KNN)}
KNN is a non-parametric, instance-based algorithm that classifies samples based on distance similarity \cite{knn-figshare, havaei2014efficient}. Using HOG features, it captures tumor shape and texture. This study evaluates Euclidean, Manhattan, Minkowski, and Chebyshev distance metrics while varying \(k\) from 1 to 10 to optimize the classification performance.

\subsubsection{Support Vector Machines (SVM)}
SVM is widely used in high-dimensional medical imaging \cite{svmkernal, Dheepak2020, sadad2021}, mapping non-linearly separable data to higher dimensions via kernel functions. The evaluation of this research is based on seven kernels: Linear, RBF, Polynomial, Sigmoid, Chi-Square, Laplacian, and Gaussian—for classifying glioma, meningioma, and pituitary tumors.

\subsubsection{CNN-MRI}
\begin{figure}[!th]
    \centering
    \fbox{\includegraphics[width=1\linewidth]{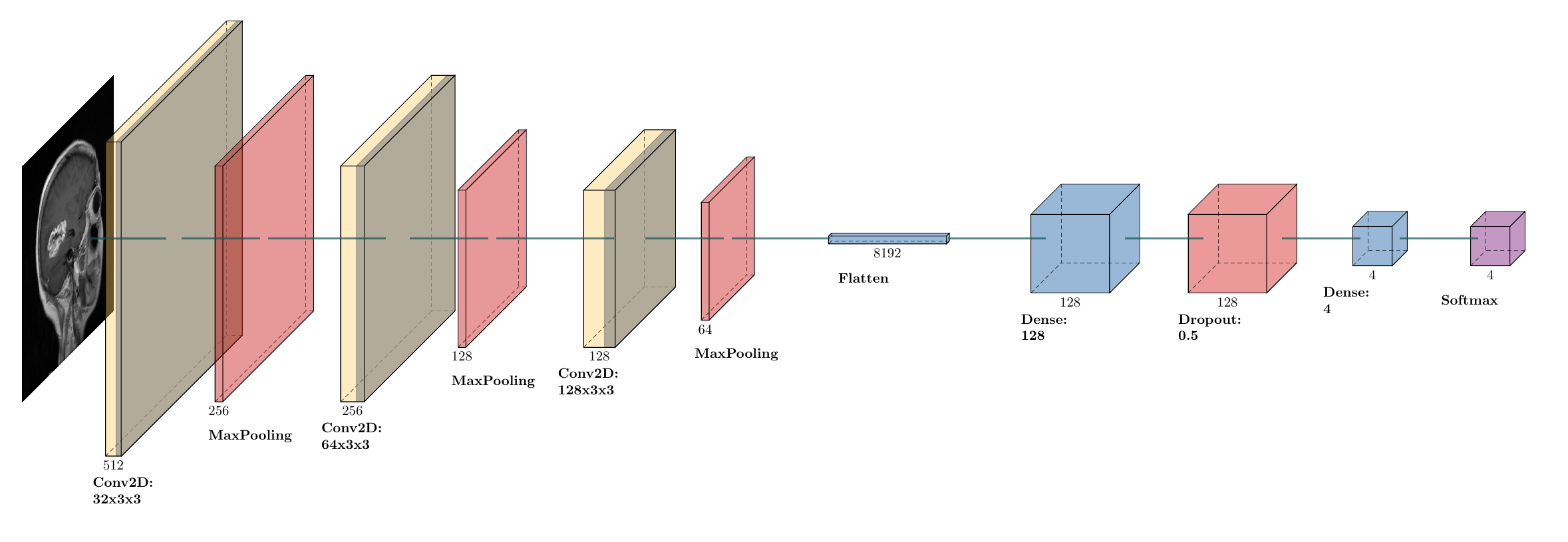}}
    \caption{The CNN-MRI architecture used in the experiments.}
    \label{fig:cnnarchitect}
\end{figure}

Convolutional Neural Networks (CNNs) extract hierarchical features directly from MRI scans \cite{Khan2020, Rasheed2023, Rasheed2024}. The CNN-MRI model (see Figure \ref{fig:cnnarchitect}) is designed to process full-resolution grayscale MRI images, which proved in our previous work \cite{Vu2024}. It consists of three convolutional layers, where lower layers focus on detecting edges and textures, while deeper layers identify tumor-specific structures. To mitigate overfitting, the network is trained using the Adam optimizer with a dropout layer (0.5 rate). The final softmax layer classifies tumors into three distinct categories.

\subsubsection{DensNet121}
DenseNet121 improves feature learning by connecting all layers, enhancing tumor structure recognition while preserving spatial details \cite{Mzoughi2023DeepTL, densnet121}. Its dense connectivity aids in detecting subtle tumor morphology differences.

\subsubsection{ResNet}
ResNet50 and ResNet101 improve feature learning by addressing the vanishing gradient problem \cite{Resnet50, Zhang2021}. ResNet50 enhances tumor boundaries using Canny edge detection, while ResNet101, with its deeper architecture, captures textures and semantic patterns for better tumor recognition.

\subsubsection{Xception}
Xception \cite{xception} improves brain tumor classification using depthwise separable convolutions, reducing complexity while preserving accuracy. It enhances boundary details in edge-enhanced MRI images and leverages pre-trained ImageNet weights to capture intricate tumor morphology, strengthening ensemble classification.

\subsection{Voting system}

\begin{figure}[!th]
    \centering
    \resizebox{\columnwidth}{!}{
    \fbox{\includegraphics[width=1\linewidth]{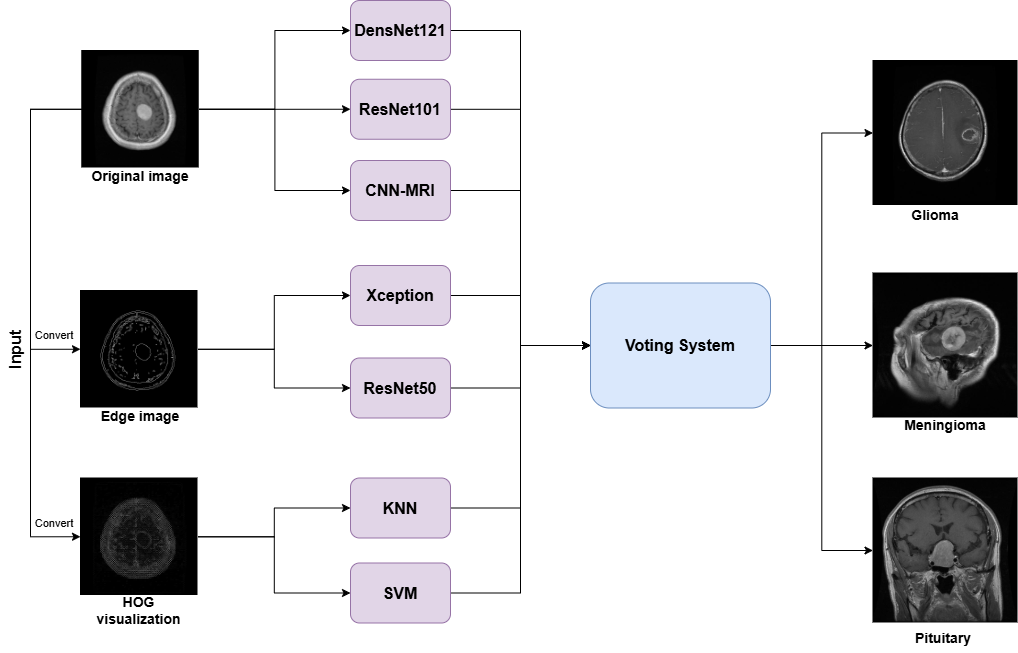}}}
    \caption{The voting system overview.}
    \label{fig:sys-overview}
\end{figure}

Figure \ref{fig:sys-overview} depicts the workflow of the voting system, where MRI images undergo preprocessing before classification. Depending on model requirements, images are used in their original form or transformed into alternative representations, such as HOG features or edge-detected images, to enhance feature extraction. Each classifier processes its respective input format and independently predicts the tumor class.

A weighted voting mechanism determines the final classification by assigning greater influence to higher accuracy models, ensuring reliable predictions. This strategy balances classifier strengths while minimizing the impact of weaker models. A mathematical formulation optimally integrates these weighted votes, selecting the class with the highest score as the final prediction.

The voting system follows the mathematical formulation:

\begin{equation}
y_{\text{final}} = \arg\max_c \left( \sum_{i=1}^{n} w_i \cdot p_{i,c} \right)
\end{equation}

Where \( y_{\text{final}} \) represents the predicted tumor class, \( w_i \) is the weight assigned to classifier \( i \), and \( p_{i,c} \) denotes the confidence score for class \( c \).

\subsection{Optimizations}

The optimization process refined key hyperparameters for brain tumor classification using AutoKeras \cite{autokeras} for CNN-MRI and Optuna \cite{optuna} for other models. Standardized settings included a learning rate of 0.0001 for stability, batch size of 32 for efficiency, and dropout (0.3–0.5) with L2 weight decay to prevent overfitting. Activation functions varied ReLU for most Sand wish for Xception. Optimization strategies included SGD with Momentum for ResNet50, Adam for ResNet101 and DenseNet121, and RMSprop for Xception. Categorical Cross-Entropy was the primary loss function, with Focal Loss used for class imbalance in ResNet101.

Uniform data augmentation techniques: flipping, rotation, zooming, brightness adjustments, and noise injection were applied to enhance generalization. These optimizations strengthened MRI-based brain tumor classification, ensuring robust performance across architectures.


\section{Results}
\label{results}
\begin{figure}[!th]
    \centering
    \begin{subfigure}[b]{0.24\linewidth}
        \centering
        \includegraphics[width=\linewidth]{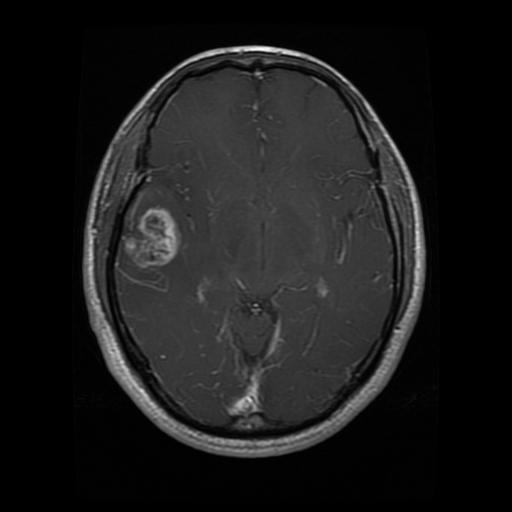}
        \caption{Glioma}
    \end{subfigure}
    \hfill
    \begin{subfigure}[b]{0.24\linewidth}
        \centering
        \includegraphics[width=\linewidth]{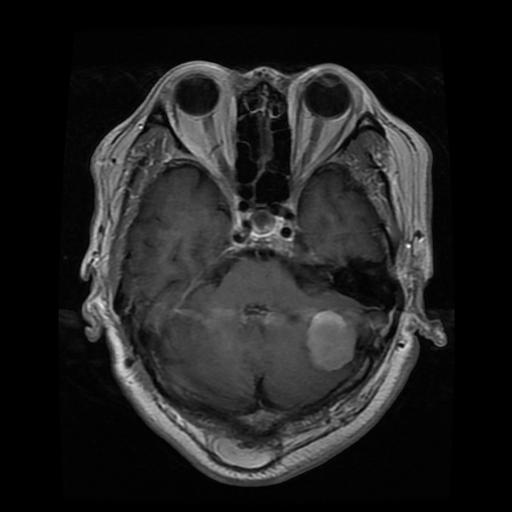}
        \caption{Meningioma}
    \end{subfigure}
    \hfill
    \begin{subfigure}[b]{0.24\linewidth}
        \centering
        \includegraphics[width=\linewidth]{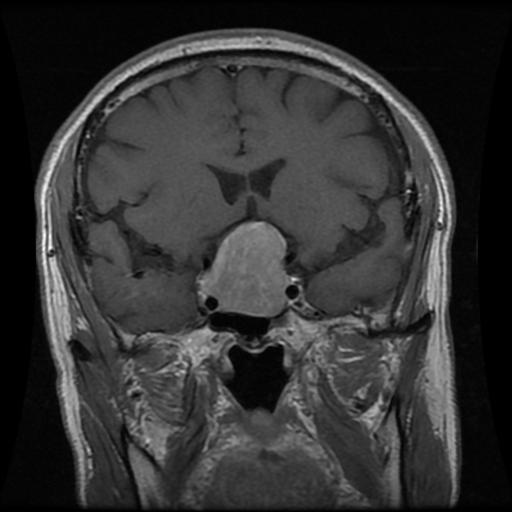}
        \caption{Pituitary}
    \end{subfigure}
    \hfill
    \begin{subfigure}[b]{0.24\linewidth}
        \centering
        \includegraphics[width=\linewidth]{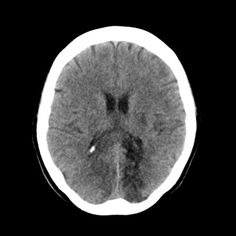}
        \caption{No Tumor}
    \end{subfigure}

    \caption{Sample images representing each tumor class: glioma, meningioma, pituitary, and no tumor.}
    \label{fig:dataset_samples}
\end{figure}

This study utilizes two well-known MRI brain tumor datasets to improve classification accuracy and generalizability: the Kaggle Brain Tumor MRI dataset \footnote{\url{https://www.kaggle.com/datasets/masoudnickparvar/brain-tumor-mri-dataset}} and the Figshare Brain Tumor dataset \footnote{\url{https://figshare.com/articles/dataset/brain_tumor_dataset/1512427}} (see Figure \ref{fig:dataset_samples}).

The \textbf{Kaggle dataset} \cite{kaggle} comprises 7,023 MRI images categorized into glioma, meningioma, pituitary tumors, and no tumor, with a predefined train-test split of 5,712 training and 1,311 testing images, ensuring balanced class distribution for consistent benchmarking. In contrast, the \textbf{Figshare dataset} contains 3,064 T1-weighted MRI images from 233 patients, divided into glioma, meningioma, and pituitary tumors with a 70\%-30\% train-test split. Unlike Kaggle, Figshare presents an imbalanced class distribution, introducing additional challenges for classification models.

\subsection{Individual Results}
Table \ref{tab:combined_results} shows that deep learning models consistently outperform traditional classifiers in accuracy and generalization. While KNN and SVM achieved competitive results, their performance depend on hyperparameters. KNN performed best with \( k = 3 \) using Euclidean distance, while SVM excelled with a linear kernel.
\begin{table}[!th]
    \centering
    \caption{Classifier performance on Figshare and Kaggle datasets.}
    \label{tab:combined_results}
    \resizebox{\columnwidth}{!}{%
    \resizebox{\textwidth}{!}{
    \begin{tabular}{|l|c|c|}
        \hline
        \textbf{Model} & \textbf{Figshare Accuracy (\%)} & \textbf{Kaggle Accuracy (\%)} \\ \hline
        KNN & 96.41 & 97.94 \\ \hline
        SVM & 95.54 & 96.03 \\ \hline
        ResNet50 & 96.63 & 98.78 \\ \hline
        Xception & 97.93 & 99.46 \\ \hline
        CNN-MRI & 95.65 & 98.32 \\ \hline
        DenseNet121 & 98.15 & 99.60 \\ \hline
        ResNet101 & 98.37 & 99.08 \\ \hline
    \end{tabular}%
    }}
\end{table}

Among deep learning models, ResNet101 and DenseNet121 demonstrated the highest performance, achieving 98.37\% and 98.15\% on Figshare and 99.08\% and 99.60\% on Kaggle, respectively. The more profound architecture of ResNet101 enhanced hierarchical feature extraction, while DenseNet121 benefited from feature reuse and efficient gradient propagation. Xception, leveraging depthwise separable convolutions, achieved 97.93\% on Figshare and 99.46\% on Kaggle, excelling in spatial feature extraction. resnet50, trained on edge-enhanced images, performed well, particularly on Kaggle, highlighting the effectiveness of boundary-aware classification.

Overall, models trained on the Kaggle dataset exhibited higher accuracy than those trained on Figshare, mainly due to Kaggle's larger dataset size and more balanced class distribution. In contrast, the imbalanced distribution in Figshare posed additional challenges, particularly affecting SVM and CNN-MRI models. CNN-MRI, while robust, showed slightly lower accuracy (95.65\%  on Figshare, 98.32\% on Kaggle) than deeper architectures, reflecting its limited feature extraction capability.

\subsection{Voting System Results Before Optimization}

All classifiers contributed equally in the no-weight scenario (see Table \ref{tab:classification_no_weight_combined}), establishing a baseline to assess the ensemble system's performance. Despite no weighting, the system achieved 99.13\% accuracy on Figshare and 99.54\% on Kaggle. These results confirm that even a simple majority voting mechanism performs effectively, providing a strong benchmark before applying weight optimizations.









\begin{table}[!th]
\centering
\caption{Classification scores for the no-weight scenario on Figshare and Kaggle datasets.}
\label{tab:classification_no_weight_combined}
\resizebox{\columnwidth}{!}{%
\begin{tabular}{|l|c|c|c|c||c|c|c|c|}
\hline
\multirow{2}{*}{\textbf{Class}} & \multicolumn{4}{c||}{\textbf{Figshare}} & \multicolumn{4}{c|}{\textbf{Kaggle}} \\ \cline{2-9}
 & \textbf{P} & \textbf{R} & \textbf{F1} & \textbf{\#} & \textbf{P} & \textbf{R} & \textbf{F1} & \textbf{\#} \\ \hline
Glioma     & 0.97  & 0.98  & 0.97  & 428  & 1.00  & 0.99  & 0.99  & 300  \\ \hline
Meningioma & 0.94  & 0.93  & 0.94  & 213  & 0.99  & 0.99  & 0.99  & 306  \\ \hline
Pituitary  & 0.98  & 0.98  & 0.99  & 279  & 0.99  & 0.99  & 1.00  & 300  \\ \hline
No tumor   & --    & --    & --    & --   & 0.99  & 1.00  & 0.99  & 405  \\ \hline
\multicolumn{9}{|c|}{\textbf{Accuracy:} Figshare = 99.13\%, Kaggle = 99.54\%} \\ \hline
\end{tabular}%
}
\end{table}

In the incremental-weight scenario (see Table \ref{tab:classification_incremental_weight_combined}), classifiers were assigned weights based on their performance, ensuring that higher-performing models had a greater influence on the ensemble's decisions. The weighting scheme ranged from $w_1=7$, $w_2=6$,  \dots,  $w_7=1$. On the Figshare dataset, this approach achieved an accuracy of 98.36\%, slightly lower than the no-weight scenario, suggesting that equal contributions from all models may be more effective in this case. However, on the Kaggle dataset, the incremental-weighting strategy maintained a high accuracy of 99.54\%, with perfect classification for the no tumor class. 


\begin{table}[!th]
\centering
\caption{Classification scores for the incremental scenario on Figshare and Kaggle datasets.}
\label{tab:classification_incremental_weight_combined}
\resizebox{\columnwidth}{!}{%
\begin{tabular}{|l|c|c|c|c||c|c|c|c|}
\hline
\multirow{2}{*}{\textbf{Class}} & \multicolumn{4}{c||}{\textbf{Figshare}} & \multicolumn{4}{c|}{\textbf{Kaggle}} \\ \cline{2-9}
 & \textbf{P} & \textbf{R} & \textbf{F1} & \textbf{\#} & \textbf{P} & \textbf{R} & \textbf{F1} & \textbf{\#} \\ \hline
Glioma     & 0.98  & 0.98  & 0.97  & 428  & 1.00  & 0.99  & 0.99  & 300  \\ \hline
Meningioma & 0.97  & 0.97  & 0.98  & 213  & 0.98  & 1.00  & 0.99  & 306  \\ \hline
Pituitary  & 0.98  & 0.98  & 0.99  & 279  & 0.99  & 0.99  & 1.00  & 300  \\ \hline
No tumor   & --    & --    & --    & --   & 1.00  & 1.00  & 1.00  & 405  \\ \hline
\multicolumn{9}{|c|}{\textbf{Accuracy:} Figshare = 98.36\%, Kaggle = 99.54\%} \\ \hline
\end{tabular}%
}
\end{table}

In the highest-weight scenario (see Table \ref{tab:classification_highest_weight_combined}), the most accurate individual model was assigned a slightly higher weight while incorporating contributions from other classifiers. On the Figshare dataset, ResNet101 achieved the highest individual accuracy (see Table \ref{tab:combined_results}), so it was given a greater influence in the ensemble (\(w_1 = 2, w_2 = 1, w_3 = 1, \dots, w_7 = 1\)), resulting in 99.13\% accuracy, matching the performance of the no-weight scenario. Meanwhile, on the Kaggle dataset, DenseNet121 demonstrated the strongest individual performance, and emphasizing its contribution led to an optimized ensemble accuracy of 99.69\%, with perfect classification across all tumor classes. 
\begin{table}[!th] 
\centering
\caption{Classification scores for the highest weight scenario on Figshare and Kaggle datasets.}
\label{tab:classification_highest_weight_combined}
\resizebox{\columnwidth}{!}{
\begin{tabular}{|l|c|c|c|c||c|c|c|c|}
\hline
\multirow{2}{*}{\textbf{Class}} & \multicolumn{4}{c||}{\textbf{Figshare}} & \multicolumn{4}{c|}{\textbf{Kaggle}} \\ \cline{2-9}
 & \textbf{P} & \textbf{R} & \textbf{F1} & \textbf{\#} & \textbf{P} & \textbf{R} & \textbf{F1} & \textbf{\#} \\ \hline
Glioma     & 0.97  & 0.98  & 0.97  & 428  & 1.00  & 0.99  & 1.00  & 300  \\ \hline
Meningioma & 0.94  & 0.93  & 0.94  & 213  & 0.99  & 1.00  & 0.99  & 306  \\ \hline
Pituitary  & 0.98  & 0.98  & 0.99  & 279  & 1.00  & 1.00  & 1.00  & 300  \\ \hline
No tumor   & --    & --    & --    & --   & 1.00  & 1.00  & 1.00  & 405  \\ \hline
\multicolumn{9}{|c|}{\textbf{Accuracy:} Figshare = 99.13\%, Kaggle = 99.69\%} \\ \hline
\end{tabular}%
}
\end{table}

\subsection{Optimized Voting System Results}
An automated approach optimized vote weights (0–10), excluding models with a weight of 0. The top three configurations achieved 99.46\% accuracy on Figshare (Table \ref{tab:figshare_optimized}) and 99.85\% on Kaggle (Table \ref{tab:kaggle_optimized}). While all reached the same accuracy, they revealed differences in misclassification distribution.

\begin{table}[!th]
    \centering
    \caption{Top three scenarios achieving 99.46\% accuracy on the Figshare dataset.}
    \label{tab:figshare_optimized}
    \resizebox{\columnwidth}{!}{%
    \begin{tabular}{|c|c|c|c|c|c|c|c|}
    \hline
         \textbf{Scenario} & \textbf{SVM} & \textbf{KNN}  & \textbf{ResNet50} & \textbf{Xception}  & \textbf{CNN-MRI}  & \textbf{DenseNet121}  & \textbf{ResNet101}  \\ \hline
         1 & 1 & 3 & 1 & 3 & 2 & 1 & 4 \\ \hline
         2 & 1 & 1 & 0 & 1 & 1 & 1 & 2 \\ \hline
         3 & 1 & 2 & 1 & 1 & 0 & 1 & 2 \\ \hline
    \end{tabular}%
    }
\end{table}
On Figshare, top configurations prioritized ResNet101, the most accurate model, with other classifiers supporting balanced classification. On Kaggle, DenseNet121 and Xception had greater influence, aligning with their superior accuracy. While overall accuracy remained identical, weight distributions affected class-specific performance, highlighting the trade-off between accuracy and classification stability.

\begin{table}[!th]
    \centering
    \caption{Top three scenarios achieved an accuracy of 99.85\% running on Kaggle dataset.}
    \label{tab:kaggle_optimized}
    \resizebox{\columnwidth}{!}{%
    \begin{tabular}{|c|c|c|c|c|c|c|c|}
    \hline
         \textbf{Scenario}& \textbf{SVM} & \textbf{KNN}  & \textbf{ResNet50} & \textbf{Xception}  & \textbf{CNN-MRI}  & \textbf{DenseNet121}  & \textbf{ResNet101}  \\ \hline
         \textbf{1}& 1 & 1 & 1 & 4 & 2 & 4 &2 \\ \hline
         \textbf{2}& 0 & 0 & 1 & 2 & 1 & 2 & 1 \\ \hline
         \textbf{3}& 0 & 0 & 1 & 3 & 1 & 2 & 1 \\ \hline
    \end{tabular}
    }
\end{table}

\subsection{Comparison}
\label{comparison}


The comparative analysis of the Figshare and Kaggle datasets highlights the accuracy gains achieved by our ensemble-based voting system. As shown in Table \ref{table:comparison1}, prior single models like the 23-layer CNN 97.80\% and ResNet50 (97.30\%) lacked generalization, while ensemble systems (e.g., AlexNet, VGG-16, ResNet50) reached 97.55\%. In contrast, our hybrid approach KNN, SVM, ResNet50, Xception, CNN-MRI, DenseNet121, ResNet101 achieved 99.46\%, demonstrating the benefits of integrating deep learning and traditional methods.

Similarly, Table \ref{table:comparison2} confirms our method’s superiority on the Kaggle dataset. While previous CNN and EfficientNet models peaked at 98.33\% and ensemble systems (GoogleNet, ShuffleNet, SVM, KNN) at 98.40\%, our optimized voting system achieved 99.85\%, ensuring improved decision-making and robustness in medical diagnostics.
\begin{table}[!th]
\centering
\caption{Performance comparison of the proposed method with existing approaches on the Figshare dataset.}
\label{table:comparison1}
\resizebox{\linewidth}{!}{%
\resizebox{\textwidth}{!}
{
\begin{tabular}{|p{2.75cm}|p{7cm}|p{1.25cm}|}
\hline
\textbf{Method} & \textbf{Single Classifiers} & \textbf{Acc. (\%)} \\ \hline

Khan et al. \cite{Khan2020}  & 23-layer CNN  & 97.80  \\ \hline
Shnaka et al. \cite{SWATI201934} & R-CNN  & 94.60  \\ \hline
Momina et al. \cite{Momina} & ResNet50 & 95.9 \\ \hline
Montoya et al. \cite{Montoya2024} & ResNet50  & 97.30  \\ \hline
Vu et al. \cite{Vu} & ResNet50  & 96.53  \\ \hline

\textbf{Method} & \textbf{Ensemble System} & \textbf{Acc. (\%)} \\ \hline

Dheepak et al. \cite{Dheepak2020} & SVM with various kernels  & 93.00  \\ \hline
Siar et al. \cite{Siar2021} & AlexNet, VGG-16, VGG-19, ResNet50 & 97.55 \\ \hline
Munira et al. \cite{Munira} & 23-layer CNN, Random Forest, SVM & 96.52 \\ \hline
Bogacsovics et al. \cite{Bogacsovics2023} & AlexNet, MobileNetv2, EfficientNet, ShuffleNetv2 & 92.00  \\ \hline
Vu et al. \cite{Vu2024} & KNN, SVM, CNN-MRI, ResNet50 & 98.36  \\ \hline

\textbf{Proposed Method}  & KNN, SVM, ResNet50, Xception, CNN-MRI, DenseNet121, ResNet101 & \textbf{99.46} \\ \hline

\end{tabular}%
}}
\end{table}

\begin{table}[!th]
\centering
\caption{Performance comparison of the proposed method with existing approaches on the Kaggle dataset.}
\label{table:comparison2}
\resizebox{\linewidth}{!}{%
\resizebox{\textwidth}{!}
{
\begin{tabular}{|p{2.75cm}|p{7cm}|p{1.25cm}|}
\hline
\textbf{Method} & \textbf{Single Classifiers} & \textbf{Acc. (\%)}  \\ \hline

Asiri et al. \cite{Asiri}  & CNN  & 94.58 \\ \hline
Ishaq et al. \cite{Ishaq2025}  & EfficientNet  & 97.40 \\ \hline
Rasheed et al. \cite{Rasheed2023}  & CNN   & 97.85 \\ \hline
Rasheed et al. \cite{Rasheed2024}  & CNN  & 98.33 \\ \hline
Ramakrishna et al. \cite{Ramakrishna2024}  & EfficientNet  & 98.00 \\ \hline

\textbf{Method} & \textbf{Ensemble System} & \textbf{Acc. (\%)}  \\ \hline

Roy et al. \cite{roy}  & SVM, Random Forest, eXtreme Gradient Boosting  & 98.15 \\ \hline
Guzmán et al. \cite{Guzman}  & ResNet50, InceptionV3, InceptionResNetV2, Xception, MobileNetV2, EfficientNetB0  & 97.12 \\ \hline
Bansal et al. \cite{Bansal2024}  & CNN and SVM  & 98.00 \\ \hline
Ali et al. \cite{ali}  & GoogleNet, ShuffleNet, NasNet-Mobile, LDA, SVM, KNN  & 98.40 \\ \hline

\textbf{Proposed Method}  & KNN, SVM, ResNet50, Xception, CNN-MRI, DenseNet121, ResNet101  & \textbf{99.85}  \\ \hline

\end{tabular}%
}
}
\end{table}

\section{Conclusion}
\label{conclusion}
This paper proposes an ensemble-based classification system that employs a weighted voting mechanism to enhance the accuracy and reliability of MRI-based brain tumor diagnosis. By integrating traditional machine learning models (KNN, SVM) with deep learning architectures (CNN-MRI, ResNet50, Xception, DenseNet121, and ResNet101), the proposed approach effectively leverages diverse feature extraction/representation techniques and classifiers. Experimental evaluations on the Figshare and Kaggle datasets confirm that this ensemble framework outperforms individual classifiers, demonstrating the advantages of model diversity in medical image analysis. The results highlight that optimizing classifier contributions through a weighted voting strategy significantly improves classification accuracy and robustness.

The optimized ensemble achieved 99.46\% accuracy on the Figshare dataset and 99.85\% on the Kaggle dataset, surpassing both standalone models and prior ensemble-based methods. Analysis of the weighting distribution emphasizes the importance of assigning greater influence to high-performing classifiers, particularly models trained on edge-enhanced images (ResNet50, Xception) and grayscale MRI scans (DenseNet121, ResNet101). However, challenges remain, particularly in distinguishing between glioma and meningioma, indicating the need for further refinement in weighting adaptation and feature extraction. Future work should explore dynamic weighting strategies and advanced feature engineering techniques to enhance classification precision across all tumor categories.

\bibliographystyle{IEEEtran}
\bibliography{source}

@article{Farhan2025,
  author    = {A. Suliman Farhan and M. Khalid and U. Manzoor},
  title     = {Brain Tumour Diagnostics and Analysis (TDA): Segmentation, Classification and Interactive Interface},
  journal   = {Computational AI in Medicine},
  year      = {2025},
  publisher = {IEEE},
}

@article{Ishaq2025,
  author    = {Ahmad Ishaq and Fath U Min Ullah and Prince Hamandawana and Da-Jung Cho and Tae-Sun Chung},
  title     = {Improved EfficientNet Architecture for Multi-Grade Brain Tumor Detection},
  journal   = {Electronics},
  volume    = {14},
  year      = {2025},
  number    = {710},
  doi       = {10.3390/electronics14040710},
  publisher = {MDPI},
}

@INPROCEEDINGS{svmkernal,
  author={Patle, Arti and Chouhan, Deepak Singh},
  booktitle={2013 International Conference on Advances in Technology and Engineering (ICATE)}, 
  title={SVM kernel functions for classification}, 
  year={2013},
  volume={},
  number={},
  pages={1-9},
  doi={10.1109/ICAdTE.2013.6524743}}

@inproceedings{sift,
  title={Object recognition from local scale-invariant features},
  author={Lowe, David G},
  booktitle={Proceedings of the seventh IEEE international conference on computer vision},
  volume={2},
  pages={1150--1157},
  year={1999},
  organization={Ieee}
}

@inproceedings{orb,
author = {Rublee, Ethan and Rabaud, Vincent and Konolige, Kurt and Bradski, Gary},
year = {2011},
month = {11},
pages = {2564-2571},
title = {ORB: an efficient alternative to SIFT or SURF},
journal = {Proceedings of the IEEE International Conference on Computer Vision},
doi = {10.1109/ICCV.2011.6126544}
}

@article{pca,
author = { Karl   Pearson   F.R.S. },
title = {LIII. On lines and planes of closest fit to systems of points in space},
journal = {The London, Edinburgh, and Dublin Philosophical Magazine and Journal of Science},
volume = {2},
number = {11},
pages = {559-572},
year  = {1901},
publisher = {Taylor & Francis},
doi = {10.1080/14786440109462720},
}

@article{bcet,
author = {LIU JIAN GUO},
title = {Balance contrast enhancement technique and its application in image colour composition},
journal = {International Journal of Remote Sensing},
volume = {12},
number = {10},
pages = {2133--2151},
year = {1991},
publisher = {Taylor \& Francis},
doi = {10.1080/01431169108955241},

}

@article{sadad2021,
author = {Sadad, Tariq and Rehman, Amjad and Munir, Asim and Saba, Tanzila and Tariq, Usman and Ayesha, Noor and Abbasi, Rashid},
title = {Brain tumor detection and multi-classification using advanced deep learning techniques},
journal = {Microscopy Research and Technique},
doi = {https://doi.org/10.1002/jemt.23688},
year = {2021}
}

@misc{autokeras,
      title={Auto-Keras: An Efficient Neural Architecture Search System}, 
      author={Haifeng Jin and Qingquan Song and Xia Hu},
      year={2019},
      eprint={1806.10282},
      archivePrefix={arXiv},
      primaryClass={cs.LG},
}

@misc{optuna,
      title={Optuna: A Next-generation Hyperparameter Optimization Framework}, 
      author={Takuya Akiba and Shotaro Sano and Toshihiko Yanase and Takeru Ohta and Masanori Koyama},
      year={2019},
      eprint={1907.10902},
      archivePrefix={arXiv},
      primaryClass={cs.LG},
}

@inproceedings{Vu2024,
  author    = {Ha Anh Vu and KC Santosh and Szilárd Vajda},
  title     = {An Expert Voting System for Brain Tumor Classification Using MRI Images},
  booktitle = {Proceedings of the Seventh International Conference on Recent Trends in Image Processing and Pattern Recognition (RTIP2R 2024)},
  year      = {2024},
  month     = {December},
  pages     = {},
  address   = {Bhopal, India},
}

@InProceedings{Vu,
author="Vu, Ha Anh
and Vajda, Szil{\'a}rd",
editor="Antonacopoulos, Apostolos
and Chaudhuri, Subhasis
and Chellappa, Rama
and Liu, Cheng-Lin
and Bhattacharya, Saumik
and Pal, Umapada",
title="Advancing Brain Tumor Diagnosis: A Hybrid Approach Using Edge Detection and Deep Learning",
booktitle="Pattern Recognition",
year="2025",
publisher="Springer Nature Switzerland",
address="Cham",
pages="226--241",
isbn="978-3-031-78201-5"
}

@article{Montoya2024,
  author = {Sebasti\'{a}n Felipe \'{A}lvarez Montoya and Alix E. Rojas and Luis Fernando Ni\~{n}o V\'{a}squez},
  title = {Classification of Brain Tumors: A Comparative Approach of Shallow and Deep Neural Networks},
  journal = {SN Computer Science},
  volume = {5},
  number = {142},
  year = {2024},
  doi = {10.1007/s42979-023-02431-7},
  url = {https://doi.org/10.1007/s42979-023-02431-7}
}

@ARTICLE{Canny,
  author={Canny, John},
  journal={IEEE Transactions on Pattern Analysis and Machine Intelligence}, 
  title={A Computational Approach to Edge Detection}, 
  year={1986},
  doi={10.1109/TPAMI.1986.4767851}}

@inproceedings{HOG,
author = {Terayama, Masahiro and Shin, Jungpil and Chang, Won-Du},
year = {2009},
month = {10},
pages = {},
title = {Object Detection using Histogram of Oriented Gradients}
}

@misc{vu2024a,
      title={Integrating Preprocessing Methods and Convolutional Neural Networks for Effective Tumor Detection in Medical Imaging}, 
      author={Ha Anh Vu},
      year={2024},
      eprint={2402.16221},
      archivePrefix={arXiv},
      primaryClass={eess.IV}
}

@misc{Figshare,
  author = {Cheng, Jun},
  title = {Brain Tumor Dataset},
  year = {2017},
  publisher = {Figshare},
  journal = {Dataset},
  doi = {10.6084/m9.figshare.1512427.v5},
  url = {https://doi.org/10.6084/m9.figshare.1512427.v5},
}

@article{Zhang2021,
  author    = {Qi Zhang},
  title     = {A novel ResNet101 model based on dense dilated convolution for image classification},
  journal   = {SN Applied Sciences},
  year      = {2021},
  doi       = {10.1007/s42452-021-04897-7},
}

@article{xception,
 author = {Benzorgat, Nawal and Xia, Kewen and Benzorgat, Mustapha Noure Eddine},
 doi = {10.7717/peerj-cs.2425},
 journal = {PeerJ Computer Science},
 title = {Enhancing brain tumor MRI classification with an ensemble of deep learning models and transformer integration},
 year = {2024}
}

@INPROCEEDINGS{Resnet50,
  author={He, Kaiming and Zhang, Xiangyu and Ren, Shaoqing and Sun, Jian},
  booktitle={2016 IEEE Conference on Computer Vision and Pattern Recognition (CVPR)}, 
  title={Deep Residual Learning for Image Recognition}, 
  year={2016},
  doi={10.1109/CVPR.2016.90}
}

@article{Khan2020,
  title={Accurate Brain Tumor Detection using Deep Convolutional Neural Network},
  author={Khan, Md. Saikat Islam and others},
  journal={Computational and Structural Biotechnology Journal 20 (2022) },
  year={2020}
}

@article{res-survey,
 author = {Debelee, Taye Girma and Kebede, Samuel Rahimeto and Schwenker, Friedhelm and Shewarega, Zemene Matewos},
 doi = {10.3390/jimaging6110121},
 journal = {Journal of Imaging},
 number = {11},
 pages = {121},
 title = {Deep Learning in Selected Cancers’ Image Analysis—A Survey},
 volume = {6},
 year = {2020}
}

@article{Dheepak2020,
  title={MEHW-SVM multi-kernel approach for improved brain tumour classification},
  author={G. Dheepak and J. Anita Christaline and D. Vaishali},
  journal={IET Image Processing},
  year={2023},
  doi={10.1049/ipr2.12345}
}

@article{Irmak2020,
  title={Multi-Classification of Brain Tumor MRI Images Using Deep Convolutional Neural Network with Fully Optimized Framework},
  author={Irmak, Emrah},
  journal={Iranian Journal of Science and Technology, Transactions of Electrical Engineering},
  volume={45},
  pages={1015--1036},
  year={2021},
  publisher={Springer},
  doi={10.1007/s40998-021-00426-9}
}

@article{cheng-brain,
    doi = {10.1371/journal.pone.0140381},
    author = {Cheng, Jun AND Huang, Wei AND Cao, Shuangliang AND Yang, Ru AND Yang, Wei AND Yun, Zhaoqiang AND Wang, Zhijian AND Feng, Qianjin},
    journal = {PLOS ONE},
    publisher = {Public Library of Science},
    title = {Enhanced Performance of Brain Tumor Classification via Tumor Region Augmentation and Partition},
    year = {2015},
    month = {10},

}

@article{SWATI201934,
title = {Brain tumor classification for MR images using transfer learning and fine-tuning},
journal = {Computerized Medical Imaging and Graphics},
year = {2019},
issn = {0895-6111},
doi = {https://doi.org/10.1016/j.compmedimag.2019.05.001},
author = {Zar Nawab Khan Swati and Qinghua Zhao and Muhammad Kabir and Farman Ali and Zakir Ali and Saeed Ahmed and Jianfeng Lu},
}

@INPROCEEDINGS{vggnet,
  author={Liu, Shuying and Deng, Weihong},
  booktitle={2015 3rd IAPR Asian Conference on Pattern Recognition (ACPR)}, 
  title={Very deep convolutional neural network based image classification using small training sample size}, 
  year={2015},
  volume={},
  number={},
  pages={730-734},
  doi={10.1109/ACPR.2015.7486599}}

@inproceedings{alexnet,
  author = {Krizhevsky, Alex and Sutskever, Ilya and Hinton, Geoffrey E},
  booktitle = {Advances in neural information processing systems},
  pages = {1097--1105},
  title = {Imagenet classification with deep convolutional neural networks},
  url = {http://papers.nips.cc/paper/4824-imagenet-classification-with-deep-convolutional-neural-networks.pdf},
  year = 2012
}

@Article{Momina,
AUTHOR = {Masood, Momina and Nazir, Tahira and Nawaz, Marriam and Mehmood, Awais and Rashid, Junaid and Kwon, Hyuk-Yoon and Mahmood, Toqeer and Hussain, Amir},
TITLE = {A Novel Deep Learning Method for Recognition and Classification of Brain Tumors from MRI Images},
JOURNAL = {Diagnostics},
PubMedID = {33919358},
ISSN = {2075-4418},
DOI = {10.3390/diagnostics11050744}
}

@Article{Abdulsalomov2023,
AUTHOR = {Abdusalomov, Akmalbek Bobomirzaevich and Mukhiddinov, Mukhriddin and Whangbo, Taeg Keun},
TITLE = {Brain Tumor Detection Based on Deep Learning Approaches and Magnetic Resonance Imaging},
JOURNAL = {Cancers},
VOLUME = {15},
YEAR = {2023},
NUMBER = {16},
ARTICLE-NUMBER = {4172},
ISSN = {2072-6694},
DOI = {10.3390/cancers15164172}
}

@misc{kaggle,
doi = {10.21227/1jny-g144},
author = {Chaki, Jyotismita},
publisher = {IEEE Dataport},
title = {Brain Tumor MRI Dataset},
year = {2023} }

@Inbook{KNN,
author="Mucherino, Antonio
and Papajorgji, Petraq J.
and Pardalos, Panos M.",
title="k-Nearest Neighbor Classification",
bookTitle="Data Mining in Agriculture",
year="2009",
publisher="Springer New York",
address="New York, NY",
isbn="978-0-387-88615-2",
doi="10.1007/978-0-387-88615-2_4",
}

@article{svm,
  title={Support-vector networks},
  author={Cortes, Corinna and Vapnik, Vladimir},
  journal={Machine learning},
  volume={20},
  number={3},
  pages={273--297},
  year={1995},
  publisher={Springer}
}

@inproceedings{Zhao2021,
  author = {X. Zhao and Y. Yuan and W. Wang and J. Chen and F. Zhao},
  title = {An Ensemble Model Based on Voting System for Brain Tumor Classification},
  booktitle = {Journal of Physics: Conference Series},
  year = {2021},
  volume = {2024},
  pages = {012010},
  doi = {10.1088/1742-6596/2024/1/012010}
}

@article{Siar2021,
  title={A combination of feature extraction methods and deep learning for brain tumour classification},
  author={Siar, Masoumeh and Teshnehlab, Mohammad},
  journal={IET Image Processing},
  volume={15},
  number={7},
  pages={1444--1456},
  year={2021},
  publisher={IET}
}

@article{Munira,
  title={Hybrid Deep Learning Models for Multi-classification of Tumour from Brain MRI},
  author={Hafiza Akter Munira and Md. Saiful Islam},
  journal={Journal of Information Systems Engineering and Business Intelligence},
  volume={8},
  number={2},
  pages={162--174},
  year={2022},
  publisher={Universitas Airlangga},
  doi={10.20473/jisebi.8.2.162-174}
}

@article{Bogacsovics2023,
  title={Developing diverse ensemble architectures for automatic brain tumor classification},
  author={Bogacsovics, Gergo and Harangi, Balazs and Hajdu, Andras},
  journal={Multimedia Tools and Applications},
  year={2023},
  publisher={Springer}
}

@article{Asiri,
  author    = {Asiri, Abdulrahman A. and Shaf, Abdul and Ali, Tariq and Aamir, Muhammad and Irfan, Muhammad and Alqahtani, Sultan},
  title     = {Enhancing brain tumor diagnosis: an optimized CNN hyperparameter model for improved accuracy and reliability},
  journal   = {PeerJ Computer Science},
  year      = {2024},
  doi       = {10.7717/peerj-cs.1878},
}

@Article{Rasheed2023,
AUTHOR = {Rasheed, Zahid and Ma, Yong-Kui and Ullah, Inam and Ghadi, Yazeed Yasin and Khan, Muhammad Zubair and Khan, Muhammad Abbas and Abdusalomov, Akmalbek and Alqahtani, Fayez and Shehata, Ahmed M.},
TITLE = {Brain Tumor Classification from MRI Using Image Enhancement and Convolutional Neural Network Techniques},
JOURNAL = {Brain Sciences},
VOLUME = {13},
YEAR = {2023},
NUMBER = {9},
ARTICLE-NUMBER = {1320},
URL = {https://www.mdpi.com/2076-3425/13/9/1320},
PubMedID = {37759920},
ISSN = {2076-3425},
DOI = {10.3390/brainsci13091320}
}

@article{Rasheed2024,
  author    = {Zahid Rasheed and Yong-Kui Ma and Inam Ullah and Mahmoud Al-Khasawneh and Sulaiman Sulmi Almutairi and Mohammed Abohashrh},
  title     = {Integrating Convolutional Neural Networks with Attention Mechanisms for Magnetic Resonance Imaging-Based Classification of Brain Tumors},
  journal   = {Bioengineering},
}

@article{Ramakrishna2024,
  author    = {Mahesh Thyluru Ramakrishna and Kuppusamy Pothanaicker and Padma Selvaraj and Surbhi Bhatia Khan and Vinoth Kumar Venkatesan and Saeed Alzahrani and Mohammad Alojail},
  title     = {Leveraging EfficientNetB3 in a Deep Learning Framework for High-Accuracy MRI Tumor Classification},
  journal   = {CMC: Computers, Materials \& Continua},
  year      = {2024},
  doi       = {10.32604/cmc.2024.053563},
}

@article{roy,
    doi = {10.1371/journal.pone.0310748},
    author = {Roy, Priyanka AND Srijon, Fahim Mohammad Sadique AND Bhowmik, Pankaj},
    journal = {PLOS ONE},
    publisher = {Public Library of Science},
    title = {An explainable ensemble approach for advanced brain tumor classification applying Dual-GAN mechanism and feature extraction techniques over highly imbalanced data},
    year = {2024},
    month = {09},

}

@Article{Guzman,
AUTHOR = {Gómez-Guzmán, Marco Antonio and Jiménez-Beristaín, Laura and García-Guerrero, Enrique Efren and López-Bonilla},
TITLE = {Classifying Brain Tumors on Magnetic Resonance Imaging by Using Convolutional Neural Networks},
JOURNAL = {Electronics},
YEAR = {2023},
ISSN = {2079-9292},
DOI = {10.3390/electronics12040955}
}

@article{Bansal2024,
title = {A Robust Hybrid Convolutional Network for Tumor Classification Using Brain MRI Image Datasets},
journal = {International Journal of Advanced Computer Science and Applications},
doi = {10.14569/IJACSA.2024.0150459},
year = {2024},
publisher = {The Science and Information Organization},
author = {Satish Bansal and Rakesh S Jadon and Sanjay K. Gupta}
}

@inproceedings{ali,
author = {Ali, Rawaa and Al-jumaili, Saif and Duru, Adil and Ucan, Osman and Boyaci, Aytug and Duru, Dilek},
year = {2022},
month = {10},
title = {Classification of Brain Tumors using MRI images based on Convolutional Neural Network and Supervised Machine Learning Algorithms},
doi = {10.1109/ISMSIT56059.2022.9932690}
}

@article{knn-figshare,
 author = {Pattanaik, Baby Barnali and Anitha, Komma and Rathore, Shanti and Biswas, Preesat and Sethy, Prabira Kumar and Behera, Santi Kumari},
 doi = {10.5114/wo.2023.124612},
 journal = {Współczesna Onkologia},
 title = {Brain tumor magnetic resonance images classification based machine learning paradigms},
 year = {2022}
}

@article{densnet121,
 author = {Yapici, Muhammed Mutlu and Karakis, Rukiye and Gurkahraman, Kali},
 doi = {10.32604/cmc.2023.035584},
 journal = {Computers Materials and Continua},
 title = {Improving Brain Tumor Classification with Deep Learning Using Synthetic Data},
 year = {2023}
}

@article{multimodels,
 author = {Sterniczuk, Bartosz and Charytanowicz, Małgorzata},
 doi = {10.12913/22998624/193627},
 journal = {Advances in Science and Technology – Research Journal},
 title = {An Ensemble Transfer Learning Model for Brain Tumors Classification using Convolutional Neural Networks},
 year = {2024}
}

@article{Mzoughi2023DeepTL,
  title={Deep Transfer Learning (DTL) Based-Framework for an Accurate Multi-classification of MRI Brain Tumors},
  author={Hiba Mzoughi and Ines Njeh and Mohamed Ben Slima and Nouha Farhat and Chokri Mhiri},
  journal={2023 International Conference on Cyberworlds (CW)},
}

@misc{densenet121,
      title={Densely Connected Convolutional Networks}, 
      author={Gao Huang and Zhuang Liu and Laurens van der Maaten and Kilian Q. Weinberger},
      year={2018},
      eprint={1608.06993},
      archivePrefix={arXiv},
      primaryClass={cs.CV},
}

@inproceedings{havaei2014efficient,
  author    = {Havaei, Mohammad and Jodoin, Pierre-Marc and Larochelle, Hugo},
  title     = {Efficient Interactive Brain Tumor Segmentation as Within-Brain kNN Classification},
  booktitle = {Proceedings of the 22nd International Conference on Pattern Recognition},
  location  = {Stockholm, Sweden},
  publisher = {IEEE},
  year      = {2014},
  month     = {August},
}

\end{document}